\documentclass[11pt]{article}

\usepackage{amsmath} 
\usepackage{palatino}
\usepackage{microtype}
\usepackage{algorithm,algpseudocode}
\usepackage{epsfig}
\usepackage{graphicx}
\usepackage[caption=false,font=footnotesize]{subfig}
\usepackage{rotating}
\def\x{\boldsymbol{x}}
\def\P{\mathbf{P}}
\begin{document}

\title{Non-Local Euclidean Medians}

\author{Kunal N. Chaudhury\thanks{Program in Applied and Computational Mathematics (PACM), Princeton University, Princeton, NJ 08544, USA (kchaudhu@math.princeton.edu). 
K. Chaudhury was partially supported by the Swiss National Science Foundation under grant PBELP2-$135867$.} \hspace{6mm} 
Amit Singer\thanks{PACM and Department of Mathematics, Princeton University, Princeton, NJ 08544, USA (amits@math.princeton.edu).
A. Singer was partially supported by Award Number DMS-0914892 from the NSF, by Award Number FA9550-09-1-0551 from AFOSR, by Award Number R01GM090200 from the National Institute of General Medical Sciences, 
and by the Alfred P. Sloan Foundation.} }

\maketitle

\begin{abstract}
In this letter, we note that the denoising performance of Non-Local Means (NLM) at large noise levels can be improved by replacing the mean by the Euclidean median.
We call this new denoising algorithm the Non-Local Euclidean Medians (NLEM). 
At the heart of NLEM is the observation that the median is more robust to outliers than the mean. 
In particular, we provide a simple geometric insight that explains why NLEM performs better than NLM in the vicinity of edges, particularly at large noise levels.  
NLEM can be efficiently implemented using iteratively reweighted least squares, and its computational complexity is comparable to that of NLM. 
We provide some preliminary results to study the proposed algorithm and to compare it with NLM.
\end{abstract}


\section{Introduction}

Non-Local Means (NLM) is a data-driven diffusion mechanism that was introduced by Buades et al. in \cite{BCM2005}. It has proved to be a simple yet powerful method for image denoising. In this method, a given pixel is denoised 
using a weighted average of other pixels in the (noisy) image. In particular, given a noisy image $u = (u_i)$, the denoised image $\hat{u} = (\hat{u}_i)$ at pixel $i$ is computed using the formula
\begin{equation}
\label{NLM_formula}
 \hat{u}_i = \frac{\sum_{j} w_{ij} u_j}{\sum_{j} w_{ij} },
\end{equation}
where $w_{ij}$ is some weight (affinity) assigned to pixels $i$ and $j$. The sum in \eqref{NLM_formula} is ideally performed over the whole image. In practice, however, one restricts $j$ to a geometric neighborhood of $i$, e.g., to a 
sufficiently large window of size $S \times S$ \cite{BCM2005}.  

The central idea in \cite{BCM2005} was to set the weights using image patches centered around each pixel, namely as 
\begin{equation}
\label{Weights}
 w_{ij} = \exp\Big(- \frac{1}{h^2}\lVert \P_i - \P_j \rVert^2  \Big),
\end{equation}
where $\P_i$ and $\P_j$ are the image patches of size $k \times k$ centered at pixels $i$ and $j$. Here, $\lVert \P \rVert$ is the Euclidean norm of patch $\P$ as a point in $\mathbf{R}^{k^2}$, and $h$ is a smoothing parameter.
Thus, pixels with similar neighborhoods are given larger weights compared to pixels whose neighborhoods look different. The algorithm makes explicit use of the fact that repetitive patterns appear in most natural images.
It is remarkable that the simple formula in \eqref{NLM_formula} often provides state-of-the-art results in image denoising \cite{BCM2010}. One outstanding feature of NLM is that, in comparison to other denoising techniques such as 
Gaussian smoothing, anisotropic diffusion, total variation denoising, and wavelet regularization, the so-called  method noise (difference of the denoised and the noisy image) in NLM appears more 
like white noise \cite{BCM2005,BCM2010}. We refer the reader to \cite{BCM2010} for a detailed review of the algorithm.

The rest of the letter is organized as follows. In Section \ref{S2}, we explain how the denoising performance of NLM can be improved in the vicinity of edges using the Euclidean median. Based on this observation, we propose a 
new denoising algorithm in Section \ref{S3} and discuss its implementation. This is followed by some preliminary denoising results in Section \ref{S4}, where we compare our new algorithm with NLM. We conclude with some remarks 
in Section \ref{S5}.

\section{Better robustness using Euclidean median}
\label{S2}

The denoising performance of NLM depends entirely on the reliability of the weights $w_{ij}$. The weights are, however, computed from the noisy image and not the clean image. Noise affects the distribution of weights, 
particularly when the noise is large. By noise, we will always mean zero-centered white Gaussian noise with variance $\sigma^2$ in the rest of the discussion. 

\begin{figure}[!htp]
  \centering
  \subfloat[Clean and noisy edge.]{\label{fig1}\includegraphics[width=0.7\linewidth]{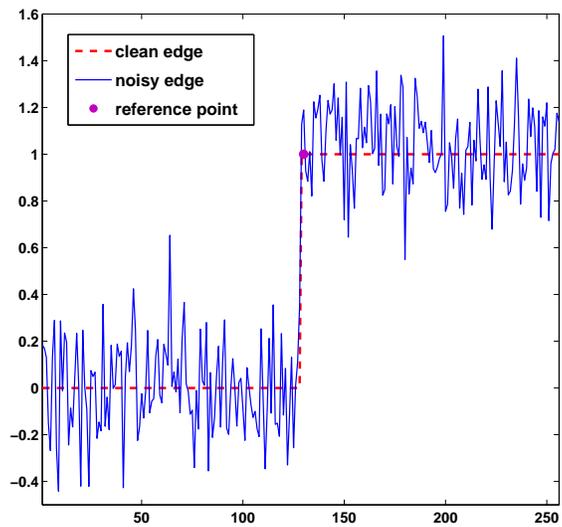}} \\
  \subfloat[Weights.]{\label{fig2}\includegraphics[width=0.7\linewidth]{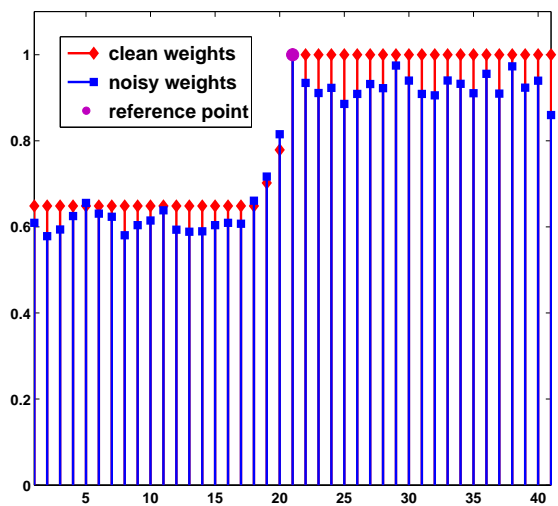}}
  \caption{Ideal edge in one dimension.}
    \label{figI}
\end{figure}

\begin{figure}[!htp]
  \centering
  \subfloat[3d patch space close to the edge.]{\label{fig3}\includegraphics[width=0.7\linewidth]{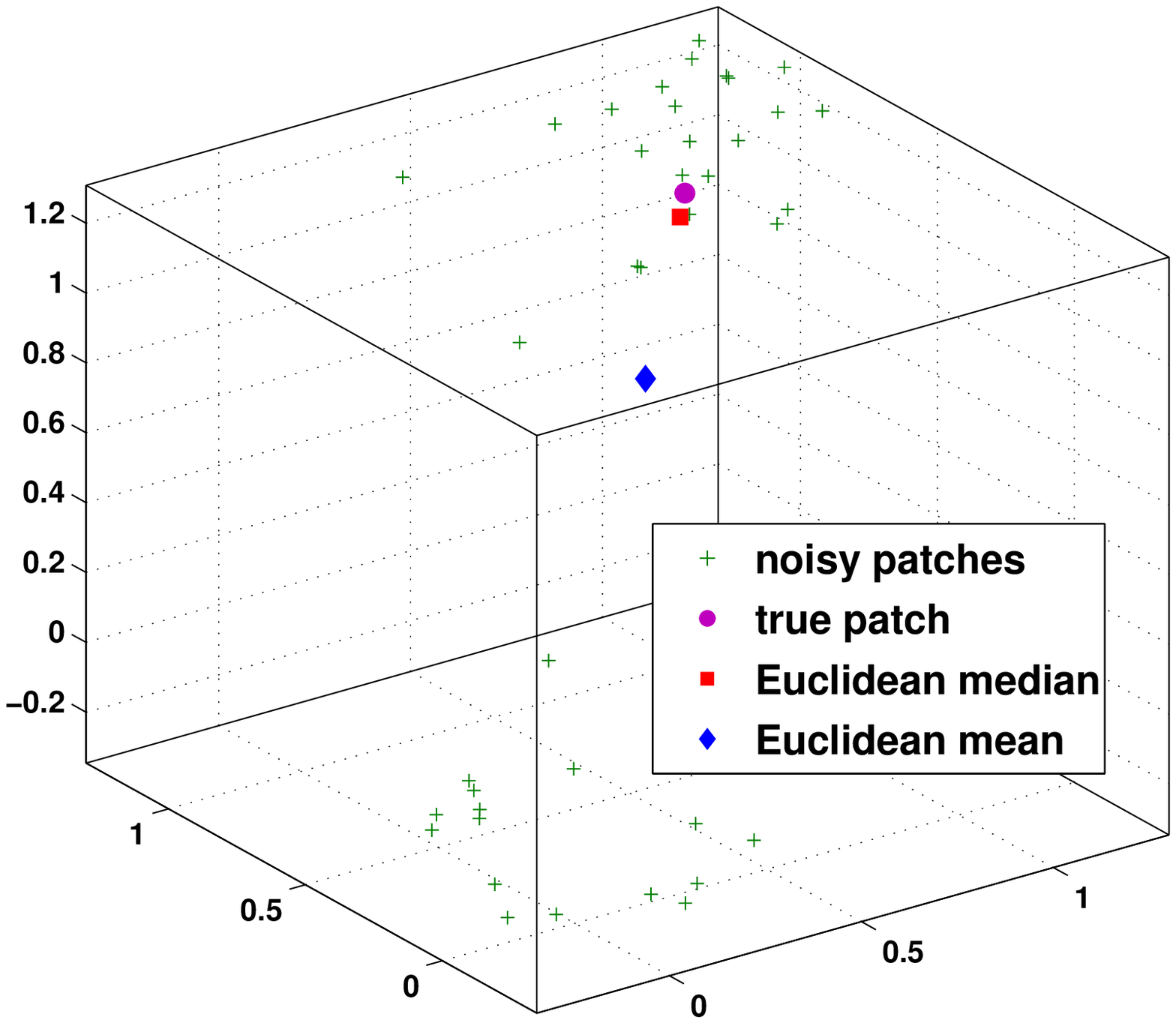}}   \\
  \subfloat[2d projection (first $2$ coordinates).]{\label{fig4}\includegraphics[width=0.7\linewidth]{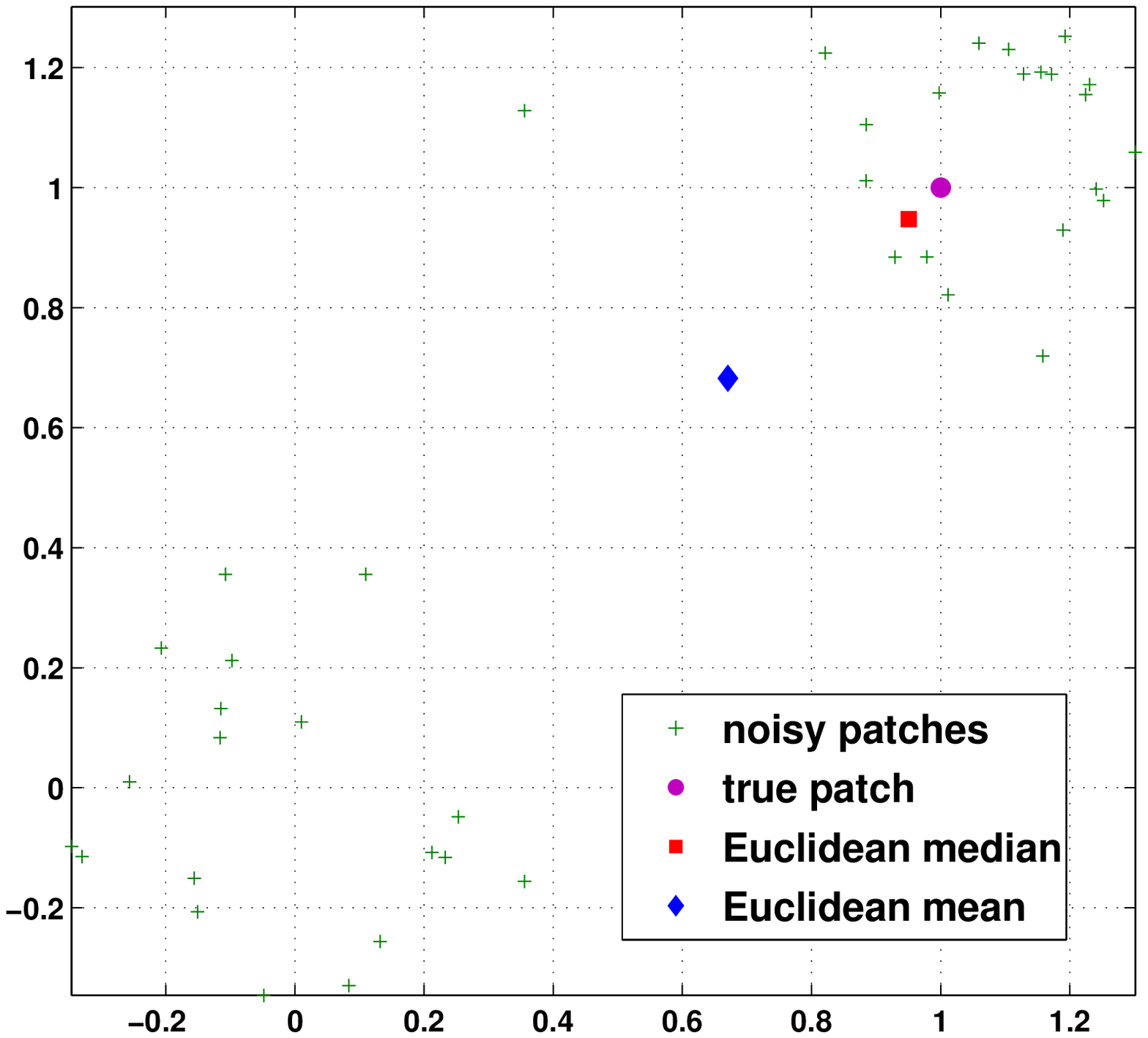}}
  \caption{Outlier model of the patch space for the point of interest in Fig. \ref{fig1}. Due to its robustness to outliers, the Euclidean median behaves as a better estimator of the clean patch than the 
  Euclidean mean (see text for explanation).}
  \label{figII}
\end{figure}

To understand the effect of noise and to motivate the main idea of the paper, we perform a simple experiment. We consider the particular case where the pixel of interest is close to an (ideal) edge. For this, we take a clean 
edge profile of unit height and add noise ($\sigma = 0.2$) to it. This is shown in Fig. \ref{figI}. We now select a point of interest a few pixels to the right of the edge (marked with a star). The goal is to estimate its true value from its neighboring points using NLM.
To do so, we take 3-sample patches around each point ($k=3$), and a search window of $S=41$. The patches are shown as points in $3$-dimensions in Fig \ref{fig3}. The clean patch for the point of interest 
is at $(1,1,1)$. 

We now use \eqref{Weights} to compute the weights, where we set $h = 10\sigma$. The weights corresponding to the point of interest are shown in Fig. \ref{fig2}. 
Using the noisy weights, we obtain an estimate of around $0.65$. This estimate has a geometric interpretation. It is the center coordinate of the Euclidean mean $\sum_{j} w_j \P_j /\sum_{j} w_j$,
where $w_j$ are the weights in Fig. \ref{fig2}, and $\P_j$ are the patches in Fig. \ref{fig3}. The Euclidean mean is marked with a purple diamond in Fig. \ref{figII}. Note that 
the patches drawn from the search window are clustered around the centers $A=(0,0,0)$ and $B=(1,1,1)$. For the point of interest, the points around $A$ are the outliers, while the ones around $B$ are the inliers.   
The noise reduces the relative gap in the corresponding weights, causing the mean to drift away from the inliers toward the outliers.

Note that the Euclidean mean is the minimizer of $\sum_{j} w_j \lVert \P - \P_j \rVert^2$ over all patches $\P$. Our main observation is that, if we instead compute the minimizer of $\sum_{j} w_j \lVert \P - \P_j \rVert$ over
all $\P$, and take the center coordinate of $\P$, then we get a much better estimate. Indeed, the denoised value turns out to be around $0.92$ in this case. 
The above minimizer is called the Euclidean median (or the geometric median) in the literature \cite{HR2009}. We will often simply call it the median. 
We repeated the above experiment using several noise realizations, and consistently got better results using the median. Averaged over $10$ trials, the denoised value using the mean and median were found to be $0.62$ and $0.93$, respectively. 
Indeed, in Fig. \ref{figII}, note how close the median is to the true patch compared to the mean. This does not come as a surprise since it is well-known that the median is more robust to outliers than the mean. This fact has 
a rather simple explanation in one dimension. In higher dimensions, note that the former is minimizer of (the square of) the weighted $\ell^2$ norm of the distances $\lVert \P - \P_j \rVert$, while the latter is the 
minimizer of the weighted $\ell^1$ norm of these distances. It is this use of the $\ell^1$ norm over the $\ell^2$ norm that makes the Euclidean median more robust to outliers \cite{HR2009}.  

\section{Non-Local Euclidean Medians}
\label{S3}

Following the above observation, we propose Algorithm \ref{algo1} below which we call the Non-Local Euclidean Medians (NLEM). We use $S(i)$ to denote the search window of size $S \times S$ centered at pixel $i$.

\begin{algorithm}
\caption{Non-Local Euclidean Medians}
\label{algo1}
\begin{algorithmic}
     \State \textbf{Input}: Noisy image $u = (u_i)$, and parameters $h, \lambda, S, k$.
      \State \textbf{Return}: Denoised image $\hat{u} = (\hat{u}_i)$.
     \State (1) Estimate noise variance $\sigma^2$, and set $h = \lambda \sigma$
     \State (2) Extract patch $\P_i \in \mathbf{R}^{k^2}$ at every pixel $i$. 
     \State (3) For every pixel $i$, do
       \State \hspace{3mm}   (a)   Set $w_{ij} = \exp(-\lVert \P_i - \P_j \rVert^2/h^2)$ for every $j \in S(i)$.
        \State \hspace{3mm}   (b)  Find patch $\P$ that minimizes $\sum_{j \in S(i)} w_{ij} \lVert \P - \P_j \rVert$.
        \State \hspace{3mm}   (c)  Assign $\hat{u}_i$ the value of the center pixel in $\P$.
\end{algorithmic}
\end{algorithm}	

The difference with NLM is in step $(3)$b which involves the computation of the Euclidean median. That is, given points $\x_1,\ldots,\x_n \in \mathbf{R}^d$ and weights $w_1,\ldots,w_n$, we are required to find $\x \in  \mathbf{R}^d$ that 
minimizes the convex cost $\sum_{j=1}^n  w_j \lVert \x - \x_j \rVert$. There exists an extensive literature on the computation of the Euclidean median; see \cite{W1937,XY1997}, and the references therein. The simplest algorithm in this area is the 
so-called Weiszfeld algorithm \cite{W1937,XY1997}. This is, in fact, based on the method of iteratively reweighted least squares (IRLS), which has received renewed interest in the compressed sensing 
community in the context of $\ell^1$  minimization \cite{CW2008,DDFG2009}. Starting from an estimate $\x^{(k)}$, the idea is to set the next iterate as 
\begin{equation*}
\x^{(k+1)} = \arg\ \min_{\x \in \mathbf{R}^d} \sum_{j=1}^n  w_j \frac{\lVert \x - \x_j \rVert^2}{\lVert \x^{(k)} - \x_j \rVert}.
\end{equation*}
This is a least-squares problem, and the minimizer is given by
\begin{equation}
\label{IRLS}
\x^{(k+1)} = \frac{\sum_j \mu^{(k)}_j \x_j}{\sum_j \mu^{(k)}_j }, 
\end{equation}
where $\mu^{(k)}_j = w_j / \lVert \x^{(k)} - \x_j \rVert$. Starting with an initial guess, one keeps repeating this process until convergence. In practice, one needs to address the situation when $\x^{(k)}$ gets close to some $\x_j$, which 
causes $\mu^{(k)}_j$ to blow up. In the Weiszfeld algorithm, one keeps track of the proximity of  $\x^{(k)}$ to all the $\x_j$, and  $\x^{(k+1)}$ is set to be $\x_i$ if $\lVert \x^{(k)} - \x_i \rVert < \varepsilon$ for some $i$. It has been proved
by many authors that the iterates converge globally, and even linearly (exponentially fast) under certain conditions, e.g., see discussion in \cite{XY1997}. 

Following the recent ideas in \cite{CW2008,DDFG2009}, we have also tried regularizing \eqref{IRLS} by adding a small bias, namely by setting $\mu^{(k)}_j = w_j / (\lVert \x^{(k)} - \x_j \rVert^2 + \varepsilon_k^2)^{1/2}$. 
The bias $\varepsilon_k$ is updated at each iterate, e.g., one starts with $\varepsilon_0 = 1$ and gradually shrinks it to zero.
The convergence properties of a particular flavor of this algorithm are discussed in \cite{DDFG2009}. We have tried both the Weiszfeld algorithm and the one in \cite{CW2008}. Experiments show us that faster 
convergence is obtained using the latter, typically within $3$ to $4$ steps. The overall computational complexity of NLEM is $k^2\cdot S^2 \cdot N_{\mathrm{iter}}$ per pixel, where $N_{\mathrm{iter}}$ is the average number of 
IRLS iterations. The complexity of the standard NLM is $k^2\cdot S^2$ per pixel.

\section{Experiments}
\label{S4}

\begin{figure}[!htp]
  \centering
  \subfloat[\textit{Checker}.]{\includegraphics[width=0.5\linewidth]{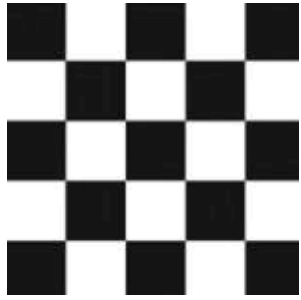}} 
  \subfloat[\textit{Circles}.]{\includegraphics[width=0.5\linewidth]{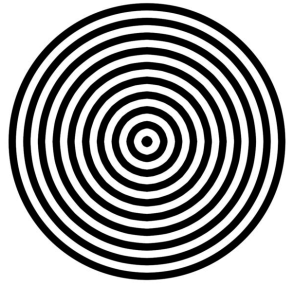}}
  \caption{Synthetic grayscale test images, original size $256 \times 256$. Black is at intensity $0$, and white is at intensity $255$.}
  \label{figIII}
\end{figure}

We now present the result of some preliminary denoising experiments. For this, we have used the synthetic images shown in Fig. \ref{figIII} and some natural images. For NLEM, we computed the Euclidean median using the simple IRLS scheme described in \cite{CW2008}. For all experiments, we have set $S = 21, k = 7$, and $h = 10 \sigma$ for both algorithms. 
These are the standard settings originally proposed in \cite{BCM2005}.

\begin{figure}[!htp]
  \centering
   \subfloat[\textit{Checker} ($\sigma$=100).]{\label{fig6}\includegraphics[width=0.5\linewidth]{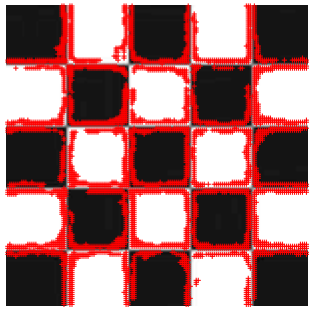}} 
  \subfloat[\textit{House} ($\sigma$=60).]{\label{fig7}\includegraphics[width=0.5\linewidth]{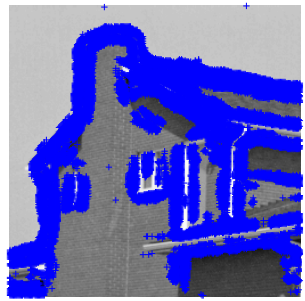}}
  \caption{The ``+'' symbol marks pixels where the estimate returned by NLEM is significantly better than that returned by NLM.}
  \label{figIV}
\end{figure}

First, we considered the \textit{Checker} image. We added noise with $\sigma =100$ resulting in a peak-signal-to-noise ratio (PSNR) of $8.18\text{dB}$. The PSNR improved to $17.94 \text{dB}$ after applying NLM, and with NLEM this
further went up to $19.45 \text{dB}$ (averaged over $10$ noise realizations).
This $1.5 \text{dB}$ improvement is perfectly explained by the arguments provided in Section \ref{S2}. Indeed, in Fig.\ref{fig6}, we have marked those pixels where the estimate from NLEM is significantly closer to the clean 
image than that obtained using NLM. More precisely, denote the original image by $f_i$, the noisy image by $u_i$, and the denoised images from NLM and NLEM by $\hat{u}_i$ and $\hat{u}'_i$. Then the ``+'' in the figure denotes 
pixels $i$ where $\lvert \hat{u}'_i - f_i \rvert < \lvert \hat{u}_i - f_i \rvert - 10$. Note that these points are concentrated around the edges where the median performs better than the mean. 

\begin{figure}[!htp]
  \centering
   \subfloat[\label{fig8}]{\includegraphics[width=0.7\linewidth]{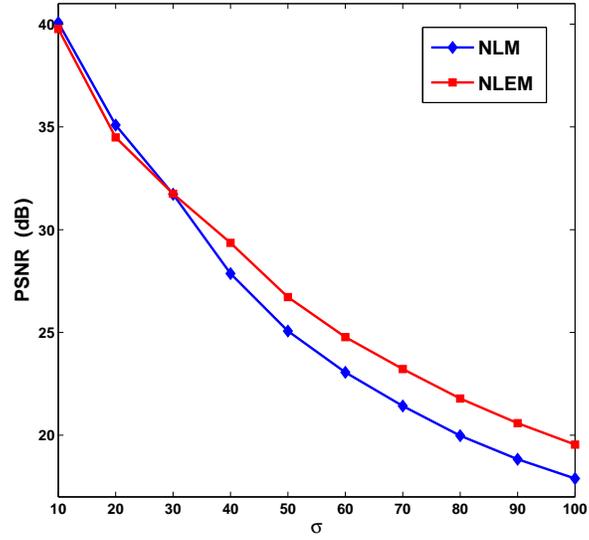}}   \\
   \subfloat[\label{fig9}]{\includegraphics[width=0.7\linewidth]{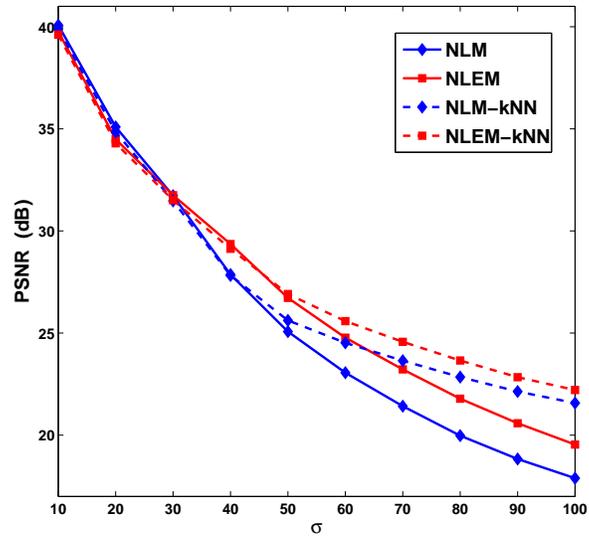}}
  \caption{Comparison of denoising performance for the \textit{Checker} image at noise $\sigma = 80$. NLM-kNN and NLEM-kNN refer to the 
  respective modifications of NLM and NLEM where we only use the top $50\%$ of the weights.}
\label{figV}
\end{figure}

So what happens when we change the noise level? We went from $\sigma = 10$ to $\sigma = 100$ in steps of $10$. The plots of the corresponding PSNRs are shown in Fig. \ref{fig8}. At low noise levels ($\sigma < 30$), we see that 
NLM performs as good or even better than NLEM. This is because at low noise levels the true neighbors in patch space are well identified, at least in the smooth regions. The difference between them is then mainly due to noise, and since the noise
is Gaussian, the least-squares approach in NLM gives statistically optimal results in the smooth regions. On the other hand, at low noise levels, the two clusters in Fig. \ref{figII} are well separated and hence the weights $w_{ij}$ for NLM are good enough to push
the mean towards the right cluster. The median and mean in this case are thus very close in the vicinity of edges. At higher noise levels, the situation completely reverses and NLEM performs consistently better than NLM. In Fig. \ref{fig8}, we see this phase 
transition occurs at around $\sigma=30$. The improvement in PSNR is quite significant beyond this noise level, ranging between $0.5 \text{dB}$ and $2.1 \text{dB}$. The exact PSNRs are given in Table \ref{table1}.

\begin{figure}[!htp]
  \centering
\subfloat{\includegraphics[width=0.5\linewidth]{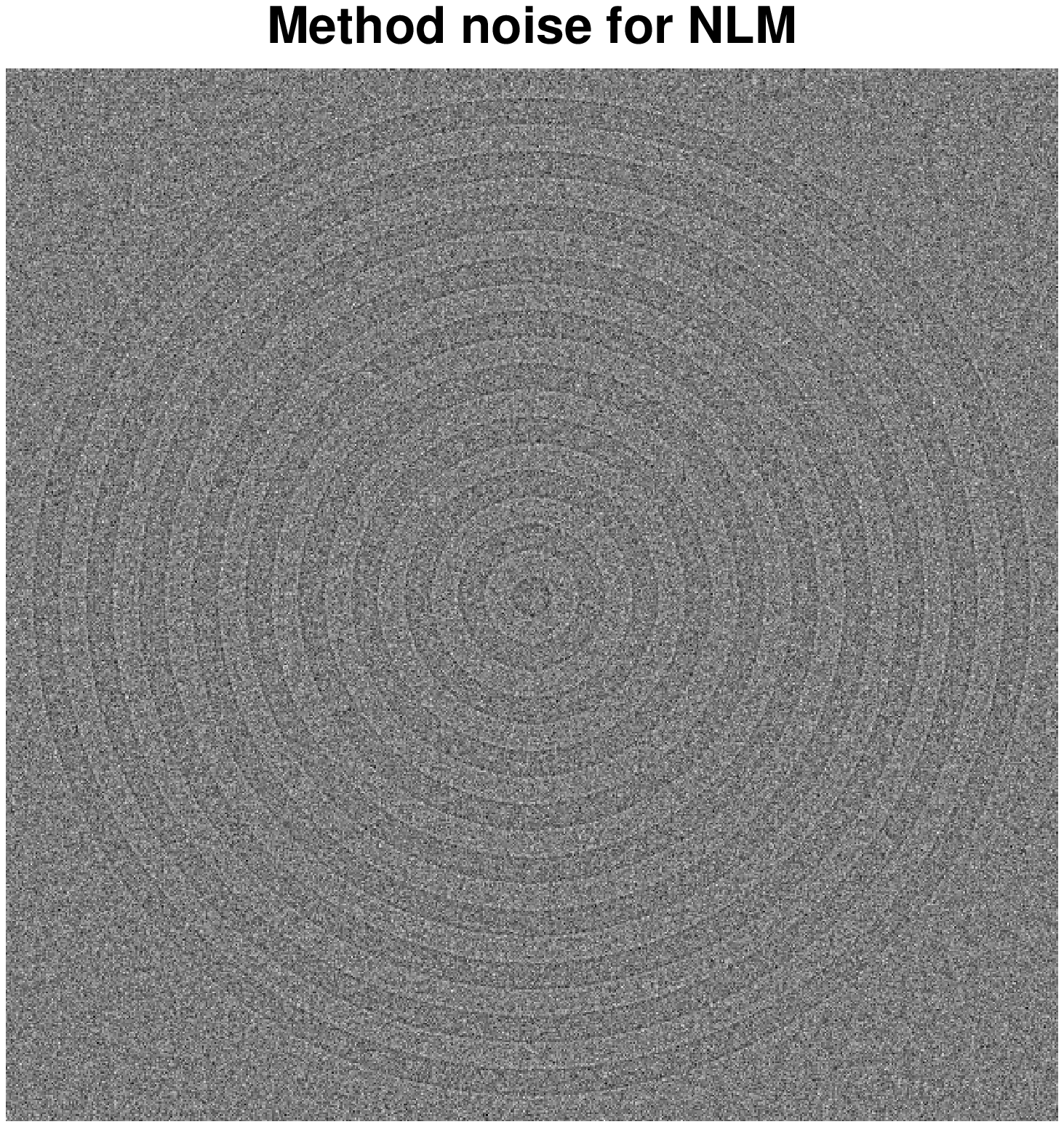}}  
{\includegraphics[width=0.5\linewidth]{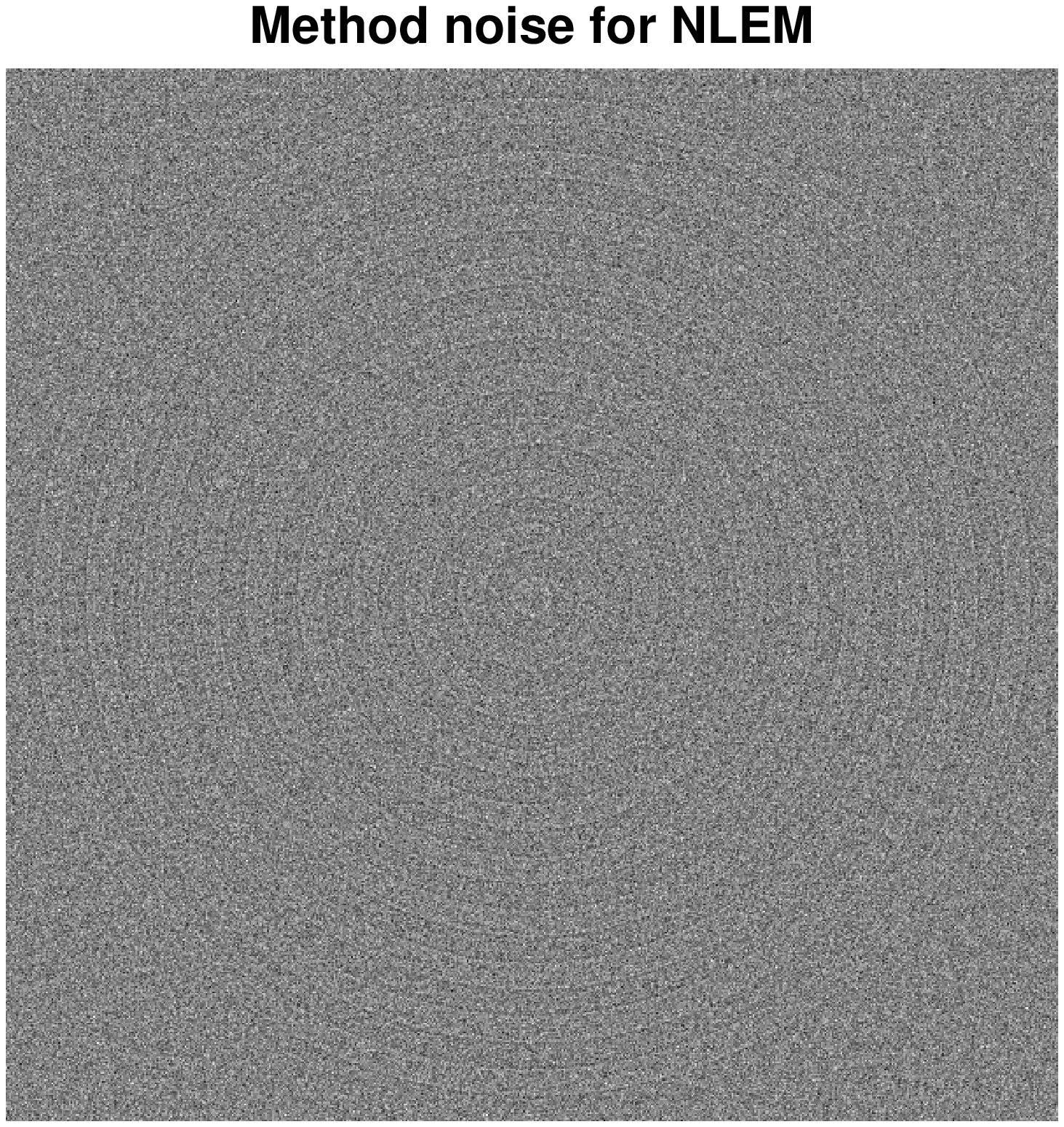}}
\caption{Comparison of the method noise for the \textit{Circles} image at $\sigma=80$.}
  \label{figVI}
\end{figure}

Next, we tried the above experiment on the \textit{Circles} image. The results are given in table \ref{table1}. The phase transition in thsi case occurs around $\sigma=25$.
NLM performs significantly better before this point, but beyond the phase transition, NLEM begins to perform better, and the gain in PSNR over NLM can be as large as $2.2\text{dB}$. 
The method noise for NLM and NLEM obtained from a typical denoising experiment are shown in Fig. \ref{figVI}. Note that the method noise for NLEM appears more white (with less structural features) than that for NLM.

Finally, we considered some benchmark natural images, namely \textit{House}, \textit{Barbara}, and \textit{Lena}. The PSNRs obtained from NLM and NLEM for these images at different noise levels are shown in Table \ref{table1}.
The table also compares the Structural SIMilarity (SSIM) indices \cite{SSIM2004} obtained from the two methods. Note that a phase transition in SSIM is also observed for some of the images. In Fig. \ref{fig7}, we show the 
pixels where NLEM does better (in the sense defined earlier) than NLM for the \textit{House} image at $\sigma = 60$. We again see that these pixels are concentrated around the edges.

The improvements in PSNR and SSIM are quite dramatic for the synthetic images \textit{Checker} and \textit{Circles}. This is expected because they contain many edges and the edges have large transitions. 
The improvement in PSNR and SSIM are less dramatic for the natural images. But, considering that NLM already provides top quality denoising results, this small improvement is already significant. We have also noticed that the performance of NLEM (and that of NLM) can be further improved by considering only the top $50\%$ of the $S^2$ neighbors with largest weights. This simple modification improves both NLM and 
NLEM (and also their run time), while still maintaining their order in terms of PSNR performance. A comparison of the four different methods is given in Fig. \ref{fig9}.
The results provided here are rather incomplete, but they already provide a good indication of the denoising potential of NLEM at large noise levels.  While we have considered a fixed setting for the parameters, our general observation based on extensive experiments is that 
NLEM consistently performs better than NLM beyond the phase transition, irrespective of the parameter settings. In future, we plan to investigate ways of further improving NLEM, and study the effect of the 
parameters on its denoising performance.

\begin{sidewaystable}
\caption{Comparison of NLM and NLEM in terms of PSNR and SSIM, at noise levels $\sigma = 10, 20, \ldots, 100$ (Averaged over $10$ noise realizations). Same parameters used: $S=21, k = 7$, and $h = 10\sigma$.}  % title name of the table
\centering  % centering table
\begin{tabular}{l  c rrrrrrrrrr}  % creating 10 columns

\hline 

\bf{Image} & \bf{Method} &\multicolumn{10}{c}{\bf{PSNR (dB)}} \\

\hline
% Checker
&NLM    &\bf{40.04}  &\bf{35.16}  &\bf{31.74}   &27.84     &25.21          &23.13       &21.39       &19.96       &18.84       &17.94 \\[-1ex]
\raisebox{1.5ex}{\textit{Checker}} 
&NLEM   &39.73       &34.66       &31.66   &\bf{29.37} &\bf{26.71}     &\bf{24.76}  &\bf{23.22}  &\bf{21.82}  &\bf{20.50}  &\bf{19.45}  \\
\hline

% Circles
&NLM    &\bf{37.31}  &\bf{34.67}       &31.79    &28.82  &26.46  & 24.69  &23.10  &21.58  &20.23  &19.03 \\[-1ex]
\raisebox{1.5ex}{\textit{Circles}} 
&NLEM   &34.27      & 34.08  &\bf{32.33} &\bf{30.05}  &\bf{27.92}   &\bf{26.24}   &\bf{24.87}  &\bf{23.57} &\bf{22.36}  &\bf{21.16} \\

\hline

% House
&NLM    &\bf{34.22}  &29.78        &26.88       &25.21        &24.07      &23.37      &22.78        &22.38      &22.06  &21.81 \\[-1ex]
\raisebox{1.5ex}{\textit{House}} 
&NLEM   &33.96       &\bf{30.10}  &\bf{27.15}  &\bf{25.39}  &\bf{24.30}  &\bf{23.51}  &\bf{22.96}  &\bf{22.54}  &\bf{22.17}  &\bf{21.95} \\

\hline

% Barbara
&NLM    &\bf{32.37}   &27.39        &24.93   &23.52       &22.64        &22.04      &21.62       &21.29      &21.07     &20.88 \\[-1ex]
\raisebox{1.5ex}{\textit{Barbara}} 
&NLEM   &32.11       &\bf{27.75} &\bf{25.26} &\bf{23.84}  &\bf{22.90}  &\bf{22.29}  &\bf{21.83}  &\bf{21.48} &\bf{21.20}  &\bf{21.01} \\

\hline

% Lena
&NLM      &\bf{33.24}   &29.31   &27.40  &26.16  &25.24  &24.54  &24.04  &23.66  &23.34  &23.06 \\[-1ex]
\raisebox{1.5ex}{\textit{Lena}} 
&NLEM     &33.15  &\bf{29.45}  &\bf{27.61}  &\bf{26.40}  &\bf{25.53}  &\bf{24.84}  & \bf{24.31}  &\bf{23.90} &\bf{23.53}  &\bf{23.24} \\
\end{tabular}

\begin{tabular}{l  c rrrrrrrrrr}  % creating 10 columns
\hline
\hline 

&  &\multicolumn{10}{c}{\bf{SSIM (\%)}} \\

\hline
% Checker
&NLM    &\bf{99.41}      &\bf{98.66}  &97.59        &95.51         &92.32           &88.35       &83.37       &78.05       &72.31       &66.47 \\[-1ex]
\raisebox{1.5ex}{\textit{Checker}} 
&NLEM   &99.36           &98.51       &\bf{97.60}   &\bf{96.30}    &\bf{ 94.03}     &\bf{91.18}  &\bf{87.81}  &\bf{84.04}  &\bf{79.59}  &\bf{74.51}  \\
\hline

% Circles
&NLM    &\bf{96.31}  &\bf{94.02}       &\bf{91.43}    &88.66        &85.79        &82.49       &78.28       &73.28       & 68.13  &63.54 \\[-1ex]
\raisebox{1.5ex}{\textit{Circles}} 
&NLEM   &95.28      &93.60             &91.20     &\bf{88.74}  &\bf{86.30}   &\bf{83.75}   &\bf{80.90}  &\bf{77.85}  &\bf{ 74.21}  &\bf{70.21} \\

\hline

% House
&NLM    &\bf{86.90}        &\bf{81.80}       &76.93        &72.97       & 69.57      &66.86       &64.38        &62.24       &60.33       &58.55 \\[-1ex]
\raisebox{1.5ex}{\textit{House}} 
&NLEM   &86.86             &81.25            &\bf{77.61}   &\bf{73.85}  &\bf{70.51}  &\bf{67.77}  &\bf{65.21}   &\bf{62.96}  &\bf{60.97}  &\bf{ 59.03} \\

\hline

% Barbara
&NLM    &89.95             &79.48        &71.32       &65.20       &60.79        &57.42      &54.64        &52.58      &50.78     &49.27 \\[-1ex]
\raisebox{1.5ex}{\textit{Barbara}} 
&NLEM   &\bf{90.25}       &\bf{80.38}    &\bf{72.49} &\bf{66.48}  &\bf{62.04}    &\bf{58.57}  &\bf{55.64}  &\bf{53.45} &\bf{51.50}  &\bf{49.86} \\

\hline

% Lena
&NLM      &86.95        &80.41      &76.45       &73.42        &70.80      &68.45        &66.46         &64.66    &62.94            &61.32 \\[-1ex]
\raisebox{1.5ex}{\textit{Lena}} 
&NLEM     &\bf{87.06}  &\bf{80.57}  &\bf{76.77}  &\bf{73.97}  &\bf{71.50}  &\bf{69.21}  & \bf{67.20}  &\bf{65.32}  &\bf{ 63.52}  &\bf{61.81} \\

\hline                          
\end{tabular}
\label{table1}
\end{sidewaystable}

\section{Discussion}
\label{S5}

The purpose of this note was to communicate the idea that one can improve (often substantially) the denoising results of NLM by replacing the $\ell^2$ regression on patch space by the more robust $\ell^1$ regression. This led us to propose the NLEM algorithm. The experimental results presented in this paper reveal two facts: (a) Phase transition phenomena -- NLEM starts to perform better (in terms of PSNR)
beyond a certain noise level, and (b) The bulk of the improvement comes from pixels close to sharp edges. The latter fact indicates that NLEM is better suited for denoising images 
that have many sharp edges. This suggests that we could get similar PSNR improvements if we simply used NLEM in the vicinity of edges and the less expensive NLM elsewhere. Unfortunately, it is difficult to get a reliable 
edge map from the noisy image. On the other hand, observation (b) suggests that, by comparing the denoising results of NLM and NLEM, one can devise a robust method of detecting edges at large noise levels. We note that Wang et al. have recently proposed a non-local median-based estimator in \cite{WSL}. This can be considered as a special case of NLEM corresponding to $k=1$, where single pixels (instead of patches) are used
for the median computation. On the other hand, some authors have proposed median-based estimators for NLM where the noisy image is median filtered before computing the weights \cite{CFFM2010,WSL}. 
In fact, most of the recent innovations in NLM were concerned with better ways of computing the weights; e.g., see \cite{T2009,deVK2011}, and reference therein. 
It would be interesting to see if the idea of robust Euclidean medians could be used to complement these innovations.

\end{document}